\documentclass[sigconf, nonacm]{acmart}

\usepackage{soul}
\usepackage{color}
\usepackage{subcaption}
\usepackage{enumitem}
\usepackage{hyperxmp}

\usepackage{tikz}

\usetikzlibrary{shapes,arrows,fit,calc,positioning}
\usetikzlibrary{arrows.meta}

\tikzset{box/.style={draw, diamond, thick, text centered, minimum height=0.5cm, minimum width=0.5cm}}
\tikzset{leaf/.style={draw, rectangle, thick, text centered, minimum height=0.25cm, minimum width=0.5cm}}
\tikzset{line/.style={draw, thick, -latex'}}

\AtBeginDocument{%
  \providecommand\BibTeX{{%
    \normalfont B\kern-0.5em{\scshape i\kern-0.25em b}\kern-0.8em\TeX}}}

\acmDOI{10.1145/nnnnnnn.nnnnnnn} % To be updated after completing copyright process
\acmISBN{978-x-xxxx-xxxx-x/YY/MM} % To be updated after completing copyright process
\acmConference[~]{~}{~}{~}
\acmYear{2022}
\copyrightyear{2022}

\hypersetup{
  pdfcopyright = {CC-BY-NC-ND 4.0},
  pdflicenseurl = {http://creativecommons.org/licenses/by-nc-nd/4.0/}
}

\newcommand\copyrighttext{%
  \footnotesize \textcopyright 2022 License: CC-BY-NC-ND 4.0 http://creativecommons.org/licenses/by-nc-nd/4.0/}
\newcommand\copyrightnotice{%
\begin{tikzpicture}[remember picture,overlay]
\node[anchor=south,yshift=10pt] at (current page.south) {\fbox{\parbox{\dimexpr\textwidth-\fboxsep-\fboxrule\relax}{\copyrighttext}}};
\end{tikzpicture}%
}

\begin{document}

\title{Interpretable AI for policy-making in pandemics}

\author{Leonardo Lucio Custode}
\email{leonardo.custode@unitn.it}
\orcid{0000-0002-1652-1690}
\affiliation{%
  \institution{University of Trento}
  \city{Trento}
  \country{Italy}
}

\author{Giovanni Iacca}
\email{giovanni.iacca@unitn.it}
\orcid{0000-0001-9723-1830}
\affiliation{%
  \institution{University of Trento}
  \city{Trento}
  \country{Italy}
}

\renewcommand{\shortauthors}{Custode and Iacca}

\begin{abstract}
Since the first wave of the COVID-19 pandemic, governments have applied restrictions in order to slow down its spreading.
However, creating such policies is hard, especially because the government needs to trade-off the spreading of the pandemic with the economic losses. For this reason, several works have applied machine learning techniques, often with the help of special-purpose simulators, to generate policies that were more effective than the ones obtained by governments.
While the performance of such approaches are promising, they suffer from a fundamental issue: since such approaches are based on black-box machine learning, their real-world applicability is limited, because these policies cannot be analyzed, nor tested, and thus they are not trustable. In this work, we employ a recently developed hybrid approach, which combines reinforcement learning with evolutionary computation, for the generation of interpretable policies for containing the pandemic. These policies, trained on an existing simulator, aim to reduce the spreading of the pandemic while minimizing the economic losses. Our results show that our approach is able to find solutions that are extremely simple, yet very powerful. In fact, our approach has significantly better performance (in simulated scenarios) than both previous work and government policies.
\end{abstract}




\maketitle
\copyrightnotice

\section{Introduction}
COVID-19 has changed the way of living of all the people on Earth.
In fact, since its outbreak, several countries adopted safety measures such as: social distancing, lockdowns, closing certain types of economic activities and others.
These safety measures were taken as a prevention mechanism to slow down the spreading of the pandemic in the world population.
However, some safety measures may have a relevant impact on the economy of a country.
For this reason, it is important to find a trade-off between spreading of the pandemic and economic losses.

To this end, some previous works focused on the use of simulators to estimate the virus propagation and economic losses given a policy \cite{kompella2020reinforcement, trott2021building}.
In this setting, a policy decides the safety measures to adopt in a given scenario.
However, while these works have been proven to be able to find policies that have better trade-offs between economic losses and pandemic spread than the ones used by governments, they do not employ interpretable AI methods.
Thus, even though such models do perform very well, their applicability is limited due to the lack of understandability.
More specifically, the main drawbacks of non-interpretable approaches for this task are: a) the fact that the trained models act as ``oracles'' and, thus, a policymaker cannot understand the rationale underlying a decision; and b) the fact that such models cannot be inspected and, thus, their behavior cannot be formally specified.

In this work, we propose interpretable models for policy-making in pandemics. These models, which combine decision trees (DTs) trained by means of grammatical evolution with Q-learning, are very simple and objectively interpretable.
Moreover, our solutions exhibit better performance w.r.t. non-interpretable state-of-the-art models.
Since our evolved policies are both more effective and more interpretable than existing black-box models, they are extremely suitable for creating policies to control the pandemic while avoiding unnecessary economic damage.

This paper is structured as follows.
The next section makes a short review of the state of the art.
Section \ref{sec:method} describes the method used to evolve interpretable DTs.
In Section \ref{sec:results} we show the experimental results and, in Section \ref{sec:interpretation}, we interpret the trees obtained. Finally, in Section \ref{sec:conclusions} we draw the conclusions of this work and suggest future work.

\section{Related work}
\label{sec:related_work}
\subsection{Policy-making for pandemics}
\label{ssec:pandemic_sim}
In \cite{kompella2020reinforcement}, the authors propose a simulator of pandemics that is able to simulate a population at a very fine granularity, modeling the activities that each agent can perform, as well as various types of economic activities.
The goal of this simulator is to test policies with the objective of minimizing simultaneously the spreading of the pandemic and the economic damages.
The authors test various handcrafted policies, policies used from a few countries, and a deep reinforcement learning based policy.
The results show that the deep reinforcement learning based policy is able to outperform both the handmade policies and the governmental ones.

In \cite{trott2021building}, the authors propose a simulator to estimate the effects on the management of closing economic activities on both the spreading of the pandemic and the economic losses.
The proposed simulator is tailored on the U.S. economy, simulating all the 51 states and a central entity that manages subsidies.

The main difference between the approaches proposed above is that in \cite{kompella2020reinforcement} the simulator acts at a very low scale, but with a high level of detail, being able to simulate aspects of everyday life, while in \cite{trott2021building} the simulator aims to perform at a very large scale with a lower level of detail.

In \cite{miikkulainen2021prediction}, the authors use a surrogate-assisted evolutionary process to optimize an agent that has to apply restrictions to avoid the spreading of the pandemic.
They make use of two neural networks: one that acts as the policy, and the other that estimates the quality of a policy, trained on real data.
While the performance of this approach is promising, the use of black-box policies such as neural network makes the real-world application of such systems difficult \cite{rudin_stop_2019}.

\subsection{Interpretable AI}
\label{ssec:interpretable_ai}

In recent years, the need for interpretable AI emerged.
Interpretable AI \cite{barredo_arrieta_explainable_2020} is defined as the branch of AI that focuses on models that are inherently interpretable by humans.
This type of models is extremely important in high-stakes and safety critical scenarios, as the ability to inspect the model and understand its behavior becomes crucial \cite{rudin_stop_2019}.

While the field of interpretable AI is making progresses, there are some sub-fields that are still seen as crucial to the development of interpretable AI.
One of these fields is interpretable reinforcement learning \cite{rudin2021interpretable}.

Lately, several works approached the problem of interpretable reinforcement learning.

Silva et al. \cite{silva_optimization_2020} make use of differentiable DTs that are trained by means of the proximal policy optimization algorithm \cite{schulman_proximal_2017}.
While this approach has shown very promising performance in the tested environments, its performance degrades significantly when transforming the differentiable DT into a regular DT.

Dhebar et al. \cite{dhebar_interpretable-ai_2020} propose an evolutionary approach for producing DTs with nonlinear splits (i.e., hyperplanes defined by conditions) for reinforcement learning tasks.
This approach is able to obtain very good performance in the tested tasks.
However, the high non-linearity of the splits hinders the interpretability of the DTs obtained.

In \cite{custode2020evolutionary}, the authors propose a two-level optimization approach that combines grammatical evolution \cite{goos_grammatical_1998} and Q-learning \cite{watkins1989learning} for evolving DTs for reinforcement learning tasks.
The approach is tested on classic reinforcement learning tasks, and the results show that the systems obtained are competitive w.r.t. non-interpretable state-of-the-art models while being very easy to interpret.
However, this approach only works in environments with discrete action spaces.

In \cite{custode2021co}, the authors extend the approach performed in \cite{custode2020evolutionary} to make it work in scenarios with continuous action spaces.
To do so, they make use of a co-evolutionary approach \cite{potter_cooperative_1994} that allows to evolve simultaneously decision tress and ``pools'' of continuous actions.

In \cite{custode2022interpretable}, the authors use genetic programming \cite{koza_genetic_1992} and CMA-ES \cite{hansen1996adapting} to evolve interpretable DTs that are able to work in RL settings with images as input.
However, the experimental results show that the proposed approach exhibits good performance only in scenarios that are not affected by noise.

\section{Method}
\label{sec:method}

\begin{figure*}
    \centering
    \resizebox{0.9\textwidth}{!}{
     \begin{tikzpicture}
         \node[text width=2.8cm, minimum width=3cm, minimum height=0.5cm, align=center, draw=black, fill=blue!20!white] (genotype) at (-3.5,0,0) { };
         \node[text width=0.3cm, minimum width=0.3cm, minimum height=0.5cm, align=center, draw=black] () at (-4.75,0,0) {1};
         \node[text width=0.3cm, minimum width=0.3cm, minimum height=0.5cm, align=center, draw=black] () at (-4.25,0,0) {8};
         \node[text width=0.3cm, minimum width=0.3cm, minimum height=0.5cm, align=center, draw=black] () at (-3.75,0,0) {2};
         \node[text width=0.3cm, minimum width=0.3cm, minimum height=0.5cm, align=center, draw=black] () at (-3.25,0,0) {16};
         \node[text width=0.3cm, minimum width=0.3cm, minimum height=0.5cm, align=center, draw=black] () at (-2.75,0,0) {9};
         \node[text width=0.3cm, minimum width=0.3cm, minimum height=0.5cm, align=center, draw=black] () at (-2.25,0,0) {\dots};
         
         \node[text width=2cm, draw=black, minimum width=2cm, minimum height=2cm, align=center, rounded corners=0.25cm, fill=green!20!white] (gen2phe) at (-0.25,0,0) {Genotype to phenotype mapping};
         
         % Decision tree %, rounded
         \node [minimum width=2cm, minimum height=1.5cm] (tree) at (2.5, 0, 0) {\ };
         \node [box] (root) at (2.35, 0.5, 0) {$c_0$};
         \node [leaf, below=-0cm of root, xshift=-0.5cm, text=white] (l) {$\cdot$};
         \node [box, below=-0cm of root, xshift=+0.5cm] (r) {$c_1$};
         \node [leaf, below=-0cm of r, xshift=-0.5cm, text=white] (rl) {$\cdot$};
         \node [leaf, below=-0cm of r, xshift=+0.5cm, text=white] (rr) {$\cdot$};
         
         \draw (root) -| (l) node [midway, above] () {\ };
         \draw (root) -| (r) node [midway, above] () {\ };
         \draw (r) -| (rl) node [midway, above] () {\ };
         \draw (r) -| (rr) node [midway, above] () {\ };
        
         \node[text width=2cm, draw=black, minimum width=2cm, minimum height=2cm, align=center, rounded corners=0.25cm, fill=orange!20!white] (ql) at (5.25,0,0) {Q-learning};
         
         %, rounded
         \node [minimum width=2cm, minimum height=1.5cm] (trainedtree) at (7.85, 0, 0) {\ };
         \node [box] (root) at (7.75, 0.5, 0) {$c_0$};
         \node [leaf, below=-0cm of root, xshift=-0.5cm] (l) {$a_0$};
         \node [box, below=-0cm of root, xshift=+0.5cm] (r) {$c_1$};
         \node [leaf, below=-0cm of r, xshift=-0.5cm] (rl) {$a_1$};
         \node [leaf, below=-0cm of r, xshift=+0.5cm] (rr) {$a_2$};
         
         \draw (root) -| (l) node [midway, above] () {\ };
         \draw (root) -| (r) node [midway, above] () {\ };
         \draw (r) -| (rl) node [midway, above] () {\ };
         \draw (r) -| (rr) node [midway, above] () {\ };
        
         % Arrows
         \draw[-Triangle] (genotype) -- (gen2phe);
         \draw[-Triangle] (gen2phe.east) -- (tree);
         \draw[-Triangle] (tree) -- (ql);
         \draw[-Triangle] (ql) -- (trainedtree);
         \end{tikzpicture}
    }
    \caption{Graphical representation of the fitness evaluation phase.}
    \label{fig:fiteval}
\end{figure*}
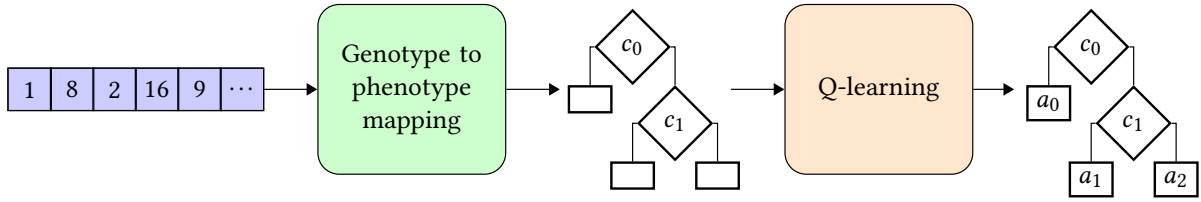

\subsection{Simulator}

Since our goal is to evolve general policies for minimizing the spread of the pandemic and, at the same time, reducing the economic losses, we employ the simulator proposed in \cite{kompella2020reinforcement}.
In fact, this simulator allows us to test rules that are applicable in every country (i.e., not only tailored to U.S. as in \cite{trott2021building}) and that do not make use of economic subsidies.

\subsubsection{State}
The simulator, at each step, provides the following features:
\begin{itemize}[leftmargin=*]
    \item $i_g$: Number of infected people since the beginning of the simulation;
    \item $r_g$: Number of recovered people since the beginning of the simulation;
    \item $c_g$: Number of patients in critical conditions since the beginning of the simulation;
    \item $d_g$: Number of dead people since the beginning of the simulation;
    \item $n_g$: Number of people that did not contract the virus from the beginning of the simulation until the current simulation day;
    \item $i_d$: Number of daily infected;
    \item $r_d$: Number of daily recovered;
    \item $c_d$: Number of patients that are in critical conditions in the current simulation day;
    \item $d_d$: Number of dead people in the current simulation day;
    \item $n_d$: Number of people that did not contract the virus in the current day\footnote{Please note that the simulator always returns $n_d=n_g$. However, for consistency with the experiments reported in \cite{kompella2020reinforcement}, in our experiments we kept both values as inputs to the policy-making agent.};
    \item $l$: The current level of ongoing restrictions;
    \item $h$: A Boolean variable indicating whether the capacity of the hospitals is saturated.
\end{itemize}
All the variables are normalized by using a min-max normalization before being fed to the policy-making agent.
Note that the simulator returns a noisy version of the estimates of the variables described above, to simulate a real-world scenario in which not all the results of the tests are known.
%(i.e., in our case, an evolved DT).

\subsubsection{Actions}
The agent can choose an action between a pool of 5 actions:
\begin{enumerate}[leftmargin=*]
    \item Stage 0: No restrictions
    \item Stage 1: Stay at home if sick, gathering limits:
    \begin{itemize}
        \item Low-risk people: 50
        \item High-risk people: 25
    \end{itemize}
    \item Stage 2: Limitations of stage 1 + wear masks + light social distancing, schools and hair salons are closed, gathering limits:
    \begin{itemize}
        \item Low-risk people: 25
        \item High-risk people: 10
    \end{itemize}
    \item Stage 3: Limitations of stage 2 + moderate social distancing + no gatherings
    \item Stage 4: Limitations of stage 3 + heavy social distancing + office and retail stores closed
\end{enumerate}
A more detailed list of actions is described in \cite{kompella2020reinforcement}.

\subsubsection{Rewards}
At each step, the agent receives a reward of:
\begin{equation}
r = -0.4 \cdot max(\frac{c_d - C}{C}, 0) -0.1 \cdot \frac{l^{1.5}}{5^{1.5}},
\end{equation}
where $C$ is the capacity of the hospitals.

Thus, the reward function aims to trade-off the number of patients in critical conditions with the stringency level of the restrictions.
Moreover, this function indirectly induces the agent to minimize also the closing of stores, limiting the economic losses.

\subsection{Evolution of decision trees}

Since the action space of this environment is discrete and the interpretability of the policies is crucial for their application in real world scenarios, we employ the method described in \cite{custode2020evolutionary} for evolving policies under the form of DTs.

Thus, we use grammatical evolution to evolve DTs with empty leaves, i.e., leaves that do not perform any action.
%in the form of ``if-then-else'' statements.
%Then, a sequence of ``if-then-else'' is converted into a DT with empty leaves, i.e., leaves that do not perform any action.
In the fitness evaluation phase, the leaves are then trained by means of Q-learning \cite{watkins1989learning}, by exploiting the reward signals coming from the environment.

A graphical illustration of the fitness evaluation phase is shown in Figure \ref{fig:fiteval}.

The evolutionary process is iterated for 50 generations using a population of 45 individuals.
All the parameters (including the ones shown in the following subsections, e.g., mutation rate and tournament size) were empirically tuned to balance performance and computational cost.
We perform 10 i.i.d. experimental runs to assess the statistical repeatability of our results.

\subsubsection{Individual encoding}
Each individual is encoded as a list of integers, ranging from 0 to a maximum value $M$, such that $M$ is significantly higher than the maximum numbers of productions in the grammar.
In our experiments, $M = 4 \cdot 10^3$.

\subsubsection{Genotype to phenotype mapping}
Each genotype (i.e., a list of integers) is transformed into its corresponding phenotype, i.e., a Python program that encodes the corresponding DT by means of ``if-then-else'' statements.
To do so, we make use of the grammar shown in Table \ref{tab:ps_ort_grammar}.
Starting from the first gene and from a production containing only the rule ``dt'', each gene is used to replace the first rule currently found in the string with the $(i~mod~k)$-th production, where $i$ is the value of the current gene and $k$ is the size of the production of the current rule (i.e., the number of choices associated to the current rule).

Note that the leaves of the resulting tree are not defined, i.e., they do not have a fixed action.
This allows us to perform Q-learning in the fitness evaluation phase, as mentioned above.

\subsubsection{Selection}
To select the individuals for mutation/crossover, we use a tournament selection with tournament size 2.

\subsubsection{Mutation}
The mutation operator used in our experiments is the uniform mutation, which replaces the current gene with a new integer uniformly sampled in $[0, M]$.
The mutation operator is applied with probability $1$ (i.e., to all the individuals in the population), while the mutation rate (e.g., the probability of mutation of each gene) is $10\%$.

\subsubsection{Crossover}
Since crossover proved to be too ``destructive'' in preliminary experiments, we do not employ it in the experiments reported below.
By destructive, we mean that, by means of this operator, the offspring can be significantly different from both parents, hindering the exploitation of good structures emerged during the evolutionary process.

\subsubsection{Replacement}
In order to increase the exploitativeness of our approach, we introduce a replacement operator.
This operator replaces a parent with an offspring only in case the offspring has (strictly) better fitness than the parent.

\subsubsection{Fitness evaluation}
The fitness of each phenotype is determined by using the corresponding policy to perform decisions in the \texttt{PandemicSimulator\footnote{\href{https://github.com/SonyAI/PandemicSimulator}{https://github.com/SonyAI/PandemicSimulator}}} environment.
To learn actions for the leaves, we employ $\varepsilon$-greedy Q-learning, with learning rate $\alpha=10^{-3}$, $\varepsilon=0.05$, and random initialization for the Q-values in $[-1, 1]$.
Moreover, we evaluate each phenotype in 10 independent episodes, each of which simulates 100 days of pandemic.

\begin{table}
    \centering
    \caption{Grammar used to produce the decision trees.}
    \label{tab:ps_ort_grammar}
    \begin{tabular}{|c|c|} \hline
        \textbf{Rule} & \textbf{Production} \\ \hline
        dt & $\langle~if\rangle~$ \\ % \hline
        if & $if\ \langle~condition\rangle~\ then\ \langle~action\rangle~\ else\ \langle~action\rangle~$ \\ % \hline
        condition & $input\_var\ \langle~comp_op\rangle~\ \langle~const_{input\_var}\rangle~$ \\ % \hline
        action & $leaf\ |\ \langle~if\rangle~$ \\ % \hline
        comp\_op & $lt\ |\ gt$ \\ % \hline
        $const_{1..10}$ & [0, 1) with step 0.1 \\ % \hline
        $const_{11,12}$ & [0, 1) with step 0.5 \\ \hline
    \end{tabular}
\end{table}

\section{Results}
\label{sec:results} 

Table \ref{tab:results_raw} shows the scores obtained by the best agents obtained in each independent run.
Here, we differentiate between \textit{training} and \textit{testing} returns.
Training returns are defined as the returns obtained by the agent in the training process, while testing refers to the returns obtained by the agent after the evolutionary process, i.e., when learning is disabled.
Moreover, we also measure the \textit{interpretability} of our solutions.
To do so, we employ the metric of interpretability proposed in \cite{virgolin_learning_2020}, reworked as done in \cite{custode2020evolutionary}.
This metric is defined as:
\[
\mathcal{M} = 0.2\ell + 0.5n_o + 3.4 n_{nao} + 4.5n_{naoc}
\]
where $\ell$ is the number of symbols appearing in the formula, $n_o$ is the number of operations contained in the model, $n_{nao}$ is the number of non-arithmetical operations, and $n_{naoc}$ is the maximum number of consecutive compositions of non-arithmetical operations.

Figure \ref{fig:comparison}, instead, compares the results obtained with the best DT produced by our method (i.e., the one corresponding to seed 3 in Table \ref{tab:results_raw}, which obtained the highest test mean return) to the policies reported in \cite{kompella2020reinforcement}. In particular, the policies used for the comparison are the following:
\begin{enumerate}[leftmargin=*]
    \item S0-4-0: Starts with stage 0, then, once 10 have been infected, it switches to stage 4 and, after 30 days, it returns to stage 0.
    \item S0-4-0FI: Similar to S0-4-0, but the return from stage 4 to stage 0 is performed gradually, by reducing the restrictions by 1 stage each 5 days.
    \item S0-4-0GI: Similar to S0-4-0, but the intermediate stages from stage 4 to stage 0 last 10 days instead of 5.
    \item S0: Always applies stage 0 restrictions.
    \item S1: Always applies stage 1 restrictions.
    \item S2: Always applies stage 2 restrictions.
    \item S3: Always applies stage 3 restrictions.
    \item S4: Always applies stage 4 restrictions.
\end{enumerate}
From Figure \ref{fig:comparison}, we can see that our best DT outperforms all the hand-crafted policies under comparison (note that the negative reward is to be maximized).
The statistical difference between the best decision tree evolved and the hand-crafted policies has been confirmed by a two-sided Wilcoxon test with $\alpha = 0.05$.

As for the deep reinforcement learning based policy presented in \cite{kompella2020reinforcement}, while we could not test it due to the fact that the model is not publicly available (and neither its numerical results are), we estimate, from the plots reported in the original paper, a cumulative reward of about -5, which is substantially lower than the cumulative reward obtained by our best DT.

Finally, in Figure \ref{fig:comparison_gov} we compare the policy obtained by means of our best DT with the ones implemented by the Italian and Swedish governments (for which we use the implementation provided in \cite{kompella2020reinforcement}). These policies act as follows:
\begin{enumerate}[leftmargin=*]
    \item ITA (approximation of the restrictions adopted by the Italian government): Increase the restrictions gradually from stage 0 to stage 4 and then it gradually returns to stage 2.
    \item SWE (approximation of restrictions adopted by the Swedish government): Applies stage 0 regulations for 3 days and then stage 1 restrictions for all the duration of the simulation.
\end{enumerate}
Also in this case, we observe that the performance are significantly better than the compared policies.
Similarly, also in this case, a Wilcoxon test with $\alpha = 0.05$ confirms the statistical significance of the results.

Moreover, we observe that the number of people infected, by using our proposed policy, is significantly smaller w.r.t. the other approaches. This fact, combined with the very high rewards obtained by our policy, suggests that the policy minimizes the economic losses by trying to stop the pandemic at the beginning, and then making the restrictions less stringent.

In the next section, we analyze the policy to understand the reasons underlying its significantly higher performance.

\begin{table}[ht!]
\centering
\caption{Results obtained by running our method in 10 i.i.d. experimental runs.}
\label{tab:results_raw}
\begin{tabular}{|l|l|l|l|}
\hline
\textbf{Seed} & \textbf{Train mean return} & \textbf{Test mean return} & \textbf{$\mathcal{M}$} \\ \hline
0  & -1.39            & -2.43 $\pm$ 1.71        & 53.40            \\
1  & -2.07            & -1.45 $\pm$ 1.36        & 17.80            \\
2  & -1.48            & -1.47 $\pm$ 1.36        & 35.60            \\
3  & -2.04            & -1.11 $\pm$ 0.32        & 35.60            \\
4  & -2.07            & -2.31 $\pm$ 1.94        & 53.40            \\
5  & -1.99            & -1.45 $\pm$ 1.34        & 35.60            \\
6  & -2.06            & -2.24 $\pm$ 1.43        & 53.40            \\
7  & -1.97            & -1.46 $\pm$ 1.37        & 53.40            \\
8  & -1.63            & -1.45 $\pm$ 1.36        & 17.80            \\
9  & -1.48            & -1.36 $\pm$ 0.90        & 53.40            \\ \hline
\end{tabular}
\end{table}

\def\imgs{cumrew_nogov, critical_nogov, dead_nogov, infected_nogov, none_nogov, recovered_nogov}

\begin{figure*}    
    \foreach \x in \imgs {
    \begin{subfigure}{0.48\textwidth}
        \resizebox{\columnwidth}{!}{
            \includegraphics[width=\columnwidth]{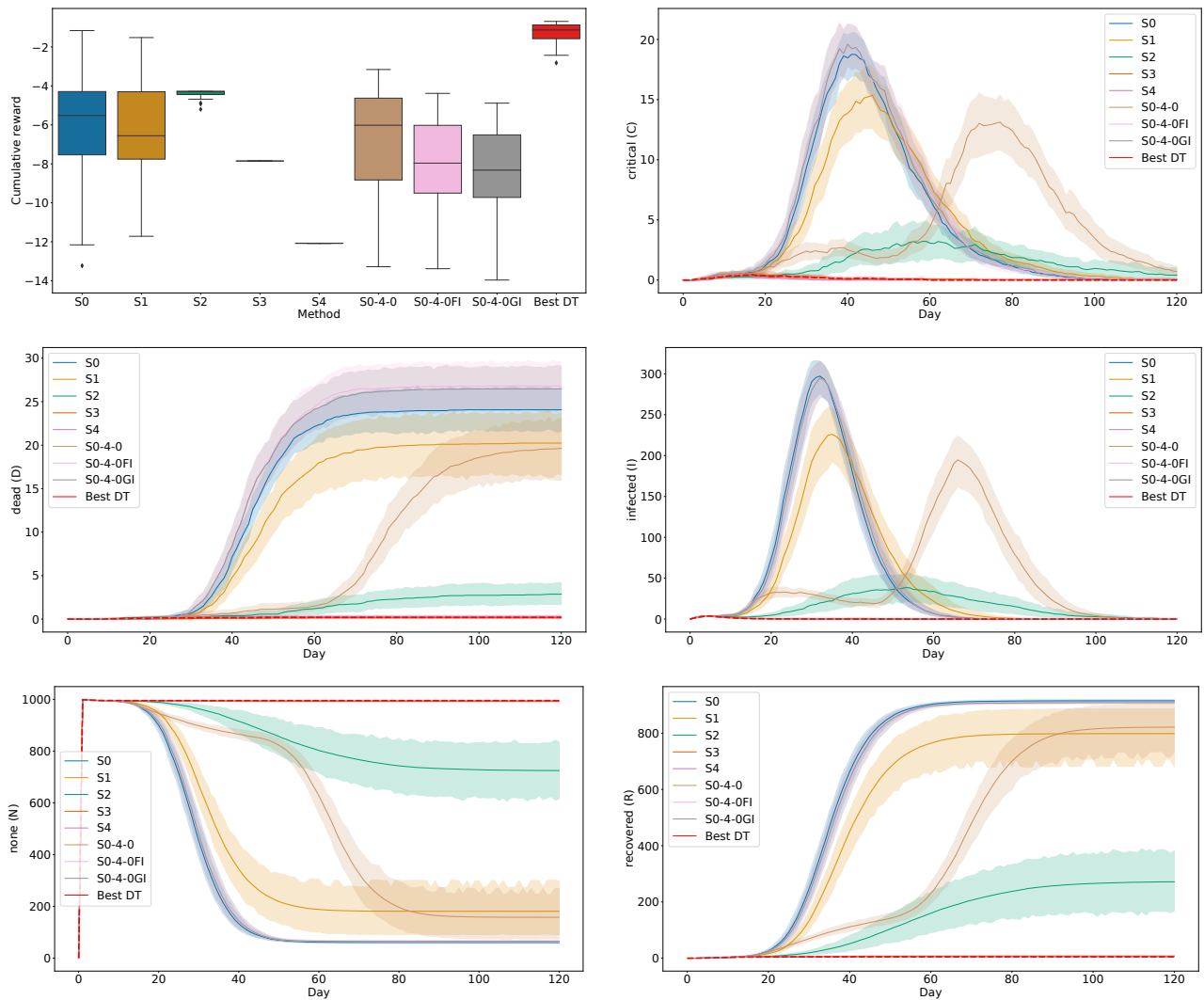}%
        }
    \end{subfigure}
    }
    \caption{Comparison of the performance of the best DT evolved w.r.t. handcrafted policies proposed in \cite{kompella2020reinforcement}.}
    \label{fig:comparison}
\end{figure*}

\def\imgs{cumrew_gov, critical_gov, dead_gov, infected_gov, none_gov, recovered_gov}

\begin{figure*}    
    \foreach \x in \imgs {
    \begin{subfigure}{0.48\textwidth}
        \resizebox{\columnwidth}{!}{
            \includegraphics[width=\columnwidth]{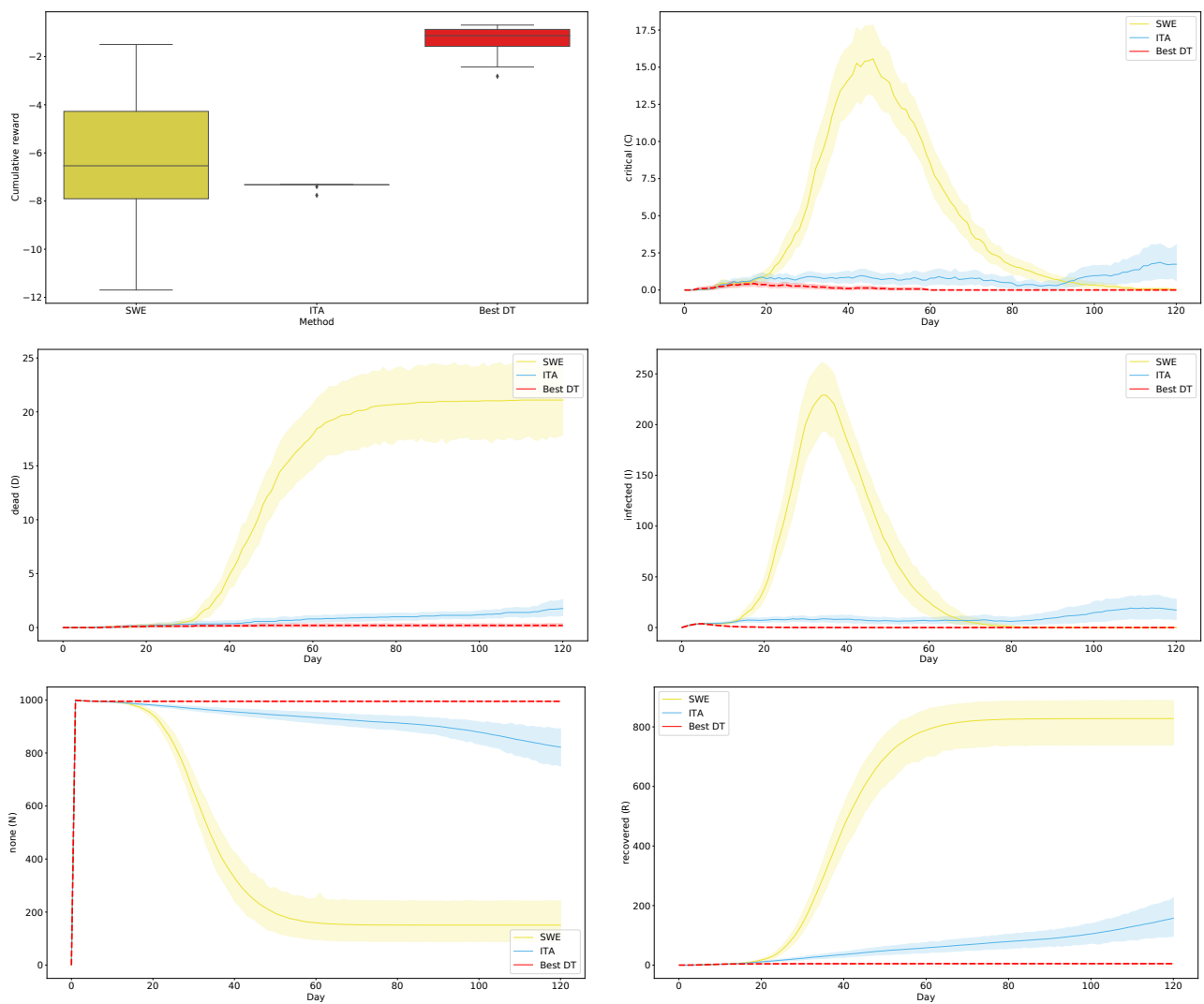}%
        }
    \end{subfigure}
    }
    \caption{Comparison of the performance of the best DT evolved w.r.t. policies adopted by the Swedish (SWE) and Italian (ITA) governments.}
    \label{fig:comparison_gov}
\end{figure*}

\section{Interpretation}
\label{sec:interpretation}
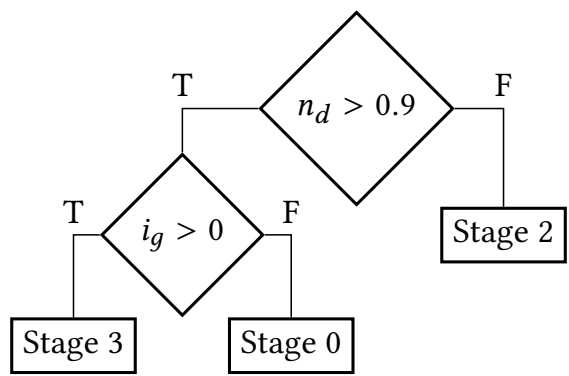
\begin{figure}
    \centering
    \resizebox{0.9\columnwidth}{!}{
    \begin{tikzpicture}[transform shape]
        \node [box] (root) {$n_d > 0.9$};
        \node [leaf, below=0.0cm of root, xshift=+1.35cm] (r) {Stage 2};
        \draw (root) -| (r) node [midway, above] () {F};
        \node [box, below=-0.5cm of root, xshift=-1.6cm] (ll) {$i_g > 0$};
        \draw (root) -| (ll) node [midway, above] () {T};
        \node [leaf, below=-0cm of ll, xshift=-1cm] (lll) {Stage 3};
        \draw (ll) -| (lll) node [midway, above] () {T};
        \node [leaf, below=-0cm of ll, xshift=+1cm] (llr) {Stage 0};
        \draw (ll) -| (llr) node [midway, above] () {F};
    \end{tikzpicture}
    }
    \caption{Best decision tree (on the test score) evolved.}
    \label{fig:best_dt}
\end{figure}

The best DT obtained is shown in Figure \ref{fig:best_dt}. While the size of the DT is quite limited, its performance are extremely satisfactory. We can see that the DT makes use of two conditions to compute its output.
%From now on, we will refer to the conditions following top-to-bottom, left-to-right order. Thus, we will refer to the root as the first condition, while we will refer to the condition on the bottom-right part of the tree as the fourth one.

The first condition (the root) checks whether the number of never-infected people ($n_d$) is greater than the 90\%.
If so, then it checks the second condition. Otherwise, it applies stage 2 restrictions, regardless of the other variables.
We may hypothesize that, in the latter case, the policy does that because, in case the number of infected people (from the beginning of the pandemic) is greater than the 10\%, then the virus may spread very quickly if the restrictions are not appropriate.

The second condition, instead, checks the number of people infected so far ($i_g$).
If the known number of infected is greater than zero, it applies stage 3 restrictions. Otherwise, it applies no restrictions.

In essence, the combination of the two branches of the root aims to induce a ``strong'' response to the initial wave of the pandemic, which is then slightly relaxed after the number of total cases increases.
Thus, the overall strategy learned by the agent is to stop the pandemic at the beginning, keeping slightly less stringent restriction after the initial wave, to avoid new waves.

\section{Conclusions}
\label{sec:conclusions}
In this paper, we leverage interpretable AI methodologies to train interpretable policies for managing a pandemic.

Despite the widely-thought trade-off between interpretability and performance, the obtained policies proved to be significantly better than handcrafted policies, deep learning policies, and government policies.

It is important to note that the results shown above have been tested in a simulated scenario and, thus, their applicability to real-world scenario must be assessed.

A limitation of the current work is the simulation scenario adopted: in our work, we simulate a population of 1000 people. The assessment of the scalability of our results is left as future work.

Other future works include: testing the proposed methodology on different simulators with different objectives, and adopting more realistic scenarios w.r.t. economical aspects (e.g., subsidies) or epidemiological aspects (e.g., the possibility of the emergence of new variants).

\bibliographystyle{ACM-Reference-Format}
\bibliography{main}


\begin{thebibliography}{18}

%%% the Web address.

\ifx \showCODEN    \undefined \def \showCODEN     #1{\unskip}     \fi
\ifx \showDOI      \undefined \def \showDOI       #1{#1}\fi
\ifx \showISBNx    \undefined \def \showISBNx     #1{\unskip}     \fi
\ifx \showISBNxiii \undefined \def \showISBNxiii  #1{\unskip}     \fi
\ifx \showISSN     \undefined \def \showISSN      #1{\unskip}     \fi
\ifx \showLCCN     \undefined \def \showLCCN      #1{\unskip}     \fi
\ifx \shownote     \undefined \def \shownote      #1{#1}          \fi
\ifx \showarticletitle \undefined \def \showarticletitle #1{#1}   \fi
\ifx \showURL      \undefined \def \showURL       {\relax}        \fi
% invisible to TeX
\providecommand\bibfield[2]{#2}
\providecommand\bibinfo[2]{#2}
\providecommand\natexlab[1]{#1}
\providecommand\showeprint[2][]{arXiv:#2}

\bibitem[\protect\citeauthoryear{Barredo~Arrieta, Díaz-Rodríguez, Del~Ser,
  Bennetot, Tabik, Barbado, Garcia, Gil-Lopez, Molina, Benjamins, Chatila, and
  Herrera}{Barredo~Arrieta et~al\mbox{.}}{2020}]%
        {barredo_arrieta_explainable_2020}
\bibfield{author}{\bibinfo{person}{Alejandro Barredo~Arrieta},
  \bibinfo{person}{Natalia Díaz-Rodríguez}, \bibinfo{person}{Javier Del~Ser},
  \bibinfo{person}{Adrien Bennetot}, \bibinfo{person}{Siham Tabik},
  \bibinfo{person}{Alberto Barbado}, \bibinfo{person}{Salvador Garcia},
  \bibinfo{person}{Sergio Gil-Lopez}, \bibinfo{person}{Daniel Molina},
  \bibinfo{person}{Richard Benjamins}, \bibinfo{person}{Raja Chatila}, {and}
  \bibinfo{person}{Francisco Herrera}.} \bibinfo{year}{2020}\natexlab{}.
\newblock \showarticletitle{Explainable {Artificial} {Intelligence} ({XAI}):
  {Concepts}, taxonomies, opportunities and challenges toward responsible
  {AI}}.
\newblock \bibinfo{journal}{\emph{Information Fusion}}  \bibinfo{volume}{58}
  (\bibinfo{date}{June} \bibinfo{year}{2020}), \bibinfo{pages}{82--115}.
\newblock
\showISSN{1566-2535}
\urldef\tempurl%
\url{https://doi.org/10.1016/j.inffus.2019.12.012}
\showDOI{\tempurl}


\bibitem[\protect\citeauthoryear{Custode and Iacca}{Custode and Iacca}{2020}]%
        {custode2020evolutionary}
\bibfield{author}{\bibinfo{person}{Leonardo~Lucio Custode} {and}
  \bibinfo{person}{Giovanni Iacca}.} \bibinfo{year}{2020}\natexlab{}.
\newblock \bibinfo{title}{Evolutionary learning of interpretable decision
  trees}.
\newblock
\newblock


\bibitem[\protect\citeauthoryear{Custode and Iacca}{Custode and Iacca}{2021}]%
        {custode2021co}
\bibfield{author}{\bibinfo{person}{Leonardo~Lucio Custode} {and}
  \bibinfo{person}{Giovanni Iacca}.} \bibinfo{year}{2021}\natexlab{}.
\newblock \showarticletitle{A co-evolutionary approach to interpretable
  reinforcement learning in environments with continuous action spaces}. In
  \bibinfo{booktitle}{\emph{Symposium Series on Computational Intelligence
  (SSCI)}}. \bibinfo{publisher}{IEEE}, \bibinfo{address}{New York, NY, USA},
  \bibinfo{pages}{1--8}.
\newblock


\bibitem[\protect\citeauthoryear{Custode and Iacca}{Custode and Iacca}{2022}]%
        {custode2022interpretable}
\bibfield{author}{\bibinfo{person}{Leonardo~Lucio Custode} {and}
  \bibinfo{person}{Giovanni Iacca}.} \bibinfo{year}{2022}\natexlab{}.
\newblock \bibinfo{title}{Interpretable pipelines with evolutionarily optimized
  modules for RL tasks with visual inputs}.
\newblock
\newblock


\bibitem[\protect\citeauthoryear{Dhebar, Deb, Nageshrao, Zhu, and Filev}{Dhebar
  et~al\mbox{.}}{2020}]%
        {dhebar_interpretable-ai_2020}
\bibfield{author}{\bibinfo{person}{Yashesh Dhebar}, \bibinfo{person}{Kalyanmoy
  Deb}, \bibinfo{person}{Subramanya Nageshrao}, \bibinfo{person}{Ling Zhu},
  {and} \bibinfo{person}{Dimitar Filev}.} \bibinfo{year}{2020}\natexlab{}.
\newblock \bibinfo{title}{Interpretable-{AI} {Policies} using {Evolutionary}
  {Nonlinear} {Decision} {Trees} for {Discrete} {Action} {Systems}}.
\newblock
\newblock
\urldef\tempurl%
\url{http://arxiv.org/abs/2009.09521}
\showURL{%
\tempurl}


\bibitem[\protect\citeauthoryear{Hansen and Ostermeier}{Hansen and
  Ostermeier}{1996}]%
        {hansen1996adapting}
\bibfield{author}{\bibinfo{person}{Nikolaus Hansen} {and}
  \bibinfo{person}{Andreas Ostermeier}.} \bibinfo{year}{1996}\natexlab{}.
\newblock \showarticletitle{Adapting arbitrary normal mutation distributions in
  evolution strategies: The covariance matrix adaptation}. In
  \bibinfo{booktitle}{\emph{IEEE International Conference on Evolutionary
  Computation}}. \bibinfo{publisher}{IEEE}, \bibinfo{address}{New York, NY,
  USA}, \bibinfo{pages}{312--317}.
\newblock


\bibitem[\protect\citeauthoryear{Kompella*, Capobianco*, Jong, Browne, Fox,
  Meyers, Wurman, and Stone}{Kompella* et~al\mbox{.}}{2020}]%
        {kompella2020reinforcement}
\bibfield{author}{\bibinfo{person}{Varun Kompella*}, \bibinfo{person}{Roberto
  Capobianco*}, \bibinfo{person}{Stacy Jong}, \bibinfo{person}{Jonathan
  Browne}, \bibinfo{person}{Spencer Fox}, \bibinfo{person}{Lauren Meyers},
  \bibinfo{person}{Peter Wurman}, {and} \bibinfo{person}{Peter Stone}.}
  \bibinfo{year}{2020}\natexlab{}.
\newblock \bibinfo{title}{Reinforcement Learning for Optimization of COVID-19
  Mitigation policies}.
\newblock
\newblock
\showeprint[arxiv]{2010.10560}~[cs.LG]


\bibitem[\protect\citeauthoryear{Koza}{Koza}{1992}]%
        {koza_genetic_1992}
\bibfield{author}{\bibinfo{person}{John~R. Koza}.}
  \bibinfo{year}{1992}\natexlab{}.
\newblock \bibinfo{booktitle}{\emph{Genetic programming: on the programming of
  computers by means of natural selection}}.
\newblock \bibinfo{publisher}{MIT Press}, \bibinfo{address}{Cambridge, Mass}.
\newblock
\showISBNx{978-0-262-11170-6}


\bibitem[\protect\citeauthoryear{Miikkulainen, Francon, Meyerson, Qiu, Sargent,
  Canzani, and Hodjat}{Miikkulainen et~al\mbox{.}}{2021}]%
        {miikkulainen2021prediction}
\bibfield{author}{\bibinfo{person}{Risto Miikkulainen},
  \bibinfo{person}{Olivier Francon}, \bibinfo{person}{Elliot Meyerson},
  \bibinfo{person}{Xin Qiu}, \bibinfo{person}{Darren Sargent},
  \bibinfo{person}{Elisa Canzani}, {and} \bibinfo{person}{Babak Hodjat}.}
  \bibinfo{year}{2021}\natexlab{}.
\newblock \showarticletitle{From prediction to prescription: evolutionary
  optimization of nonpharmaceutical interventions in the COVID-19 pandemic}.
\newblock \bibinfo{journal}{\emph{IEEE Transactions on Evolutionary
  Computation}} \bibinfo{volume}{25}, \bibinfo{number}{2}
  (\bibinfo{year}{2021}), \bibinfo{pages}{386--401}.
\newblock


\bibitem[\protect\citeauthoryear{Potter and De~Jong}{Potter and
  De~Jong}{1994}]%
        {potter_cooperative_1994}
\bibfield{author}{\bibinfo{person}{Mitchell~A. Potter} {and}
  \bibinfo{person}{Kenneth~A. De~Jong}.} \bibinfo{year}{1994}\natexlab{}.
\newblock \showarticletitle{A cooperative coevolutionary approach to function
  optimization}. In \bibinfo{booktitle}{\emph{Parallel {Problem} {Solving} from
  {Nature} — {PPSN} {III}}}. \bibinfo{publisher}{Springer},
  \bibinfo{address}{Berlin, Heidelberg}, \bibinfo{pages}{249--257}.
\newblock
\showISBNx{978-3-540-49001-2}
\urldef\tempurl%
\url{https://doi.org/10.1007/3-540-58484-6_269}
\showDOI{\tempurl}


\bibitem[\protect\citeauthoryear{Rudin}{Rudin}{2019}]%
        {rudin_stop_2019}
\bibfield{author}{\bibinfo{person}{Cynthia Rudin}.}
  \bibinfo{year}{2019}\natexlab{}.
\newblock \showarticletitle{Stop explaining black box machine learning models
  for high stakes decisions and use interpretable models instead}.
\newblock \bibinfo{journal}{\emph{Nature Machine Intelligence}}
  \bibinfo{volume}{1}, \bibinfo{number}{5} (\bibinfo{date}{May}
  \bibinfo{year}{2019}), \bibinfo{pages}{206--215}.
\newblock
\showISSN{2522-5839}
\urldef\tempurl%
\url{https://doi.org/10.1038/s42256-019-0048-x}
\showDOI{\tempurl}


\bibitem[\protect\citeauthoryear{Rudin, Chen, Chen, Huang, Semenova, and
  Zhong}{Rudin et~al\mbox{.}}{2021}]%
        {rudin2021interpretable}
\bibfield{author}{\bibinfo{person}{Cynthia Rudin}, \bibinfo{person}{Chaofan
  Chen}, \bibinfo{person}{Zhi Chen}, \bibinfo{person}{Haiyang Huang},
  \bibinfo{person}{Lesia Semenova}, {and} \bibinfo{person}{Chudi Zhong}.}
  \bibinfo{year}{2021}\natexlab{}.
\newblock \bibinfo{title}{Interpretable Machine Learning: Fundamental
  Principles and 10 Grand Challenges}.
\newblock
\newblock


\bibitem[\protect\citeauthoryear{Ryan, Collins, and Neill}{Ryan
  et~al\mbox{.}}{1998}]%
        {goos_grammatical_1998}
\bibfield{author}{\bibinfo{person}{Conor Ryan}, \bibinfo{person}{Jj Collins},
  {and} \bibinfo{person}{Michael~O Neill}.} \bibinfo{year}{1998}\natexlab{}.
\newblock \showarticletitle{Grammatical evolution: {Evolving} programs for an
  arbitrary language}.
\newblock In \bibinfo{booktitle}{\emph{{European Conference on Genetic
  Programming}}}. \bibinfo{publisher}{Springer}, \bibinfo{address}{Berlin,
  Heidelberg}, \bibinfo{pages}{83--96}.
\newblock
\showISBNx{978-3-540-64360-9 978-3-540-69758-9}
\urldef\tempurl%
\url{https://doi.org/10.1007/BFb0055930}
\showDOI{\tempurl}


\bibitem[\protect\citeauthoryear{Schulman, Wolski, Dhariwal, Radford, and
  Klimov}{Schulman et~al\mbox{.}}{2017}]%
        {schulman_proximal_2017}
\bibfield{author}{\bibinfo{person}{John Schulman}, \bibinfo{person}{Filip
  Wolski}, \bibinfo{person}{Prafulla Dhariwal}, \bibinfo{person}{Alec Radford},
  {and} \bibinfo{person}{Oleg Klimov}.} \bibinfo{year}{2017}\natexlab{}.
\newblock \bibinfo{title}{Proximal {Policy} {Optimization} {Algorithms}}.
\newblock
\newblock
\urldef\tempurl%
\url{http://arxiv.org/abs/1707.06347}
\showURL{%
\tempurl}
\newblock
\shownote{arXiv:1707.06347.}


\bibitem[\protect\citeauthoryear{Silva, Gombolay, Killian, Jimenez, and
  Son}{Silva et~al\mbox{.}}{2020}]%
        {silva_optimization_2020}
\bibfield{author}{\bibinfo{person}{Andrew Silva}, \bibinfo{person}{Matthew
  Gombolay}, \bibinfo{person}{Taylor Killian}, \bibinfo{person}{Ivan Jimenez},
  {and} \bibinfo{person}{Sung-Hyun Son}.} \bibinfo{year}{2020}\natexlab{}.
\newblock \showarticletitle{Optimization {Methods} for {Interpretable}
  {Differentiable} {Decision} {Trees} {Applied} to {Reinforcement} {Learning}}.
  In \bibinfo{booktitle}{\emph{International {Conference} on {Artificial}
  {Intelligence} and {Statistics}}}. \bibinfo{publisher}{PMLR},
  \bibinfo{address}{Palermo, Italy}, \bibinfo{pages}{1855--1865}.
\newblock
\showISSN{2640-3498}
\urldef\tempurl%
\url{http://proceedings.mlr.press/v108/silva20a.html}
\showURL{%
\tempurl}


\bibitem[\protect\citeauthoryear{Trott, Srinivasa, van~der Wal, Haneuse, and
  Zheng}{Trott et~al\mbox{.}}{2021}]%
        {trott2021building}
\bibfield{author}{\bibinfo{person}{Alexander Trott}, \bibinfo{person}{Sunil
  Srinivasa}, \bibinfo{person}{Douwe van~der Wal}, \bibinfo{person}{Sebastien
  Haneuse}, {and} \bibinfo{person}{Stephan Zheng}.}
  \bibinfo{year}{2021}\natexlab{}.
\newblock \bibinfo{title}{Building a foundation for data-driven, interpretable,
  and robust policy design using the ai economist}.
\newblock
\newblock


\bibitem[\protect\citeauthoryear{Virgolin, De~Lorenzo, Medvet, and
  Randone}{Virgolin et~al\mbox{.}}{2020}]%
        {virgolin_learning_2020}
\bibfield{author}{\bibinfo{person}{Marco Virgolin}, \bibinfo{person}{Andrea
  De~Lorenzo}, \bibinfo{person}{Eric Medvet}, {and} \bibinfo{person}{Francesca
  Randone}.} \bibinfo{year}{2020}\natexlab{}.
\newblock \showarticletitle{Learning a Formula of Interpretability to Learn
  Interpretable Formulas}. In \bibinfo{booktitle}{\emph{Parallel Problem
  Solving from Nature -- PPSN XVI}}, \bibfield{editor}{\bibinfo{person}{Thomas
  B{\"a}ck}, \bibinfo{person}{Mike Preuss}, \bibinfo{person}{Andr{\'e} Deutz},
  \bibinfo{person}{Hao Wang}, \bibinfo{person}{Carola Doerr},
  \bibinfo{person}{Michael Emmerich}, {and} \bibinfo{person}{Heike Trautmann}}
  (Eds.). \bibinfo{publisher}{Springer International Publishing},
  \bibinfo{address}{Cham}, \bibinfo{pages}{79--93}.
\newblock
\showISBNx{978-3-030-58115-2}


\bibitem[\protect\citeauthoryear{Watkins}{Watkins}{1989}]%
        {watkins1989learning}
\bibfield{author}{\bibinfo{person}{Christopher John Cornish~Hellaby Watkins}.}
  \bibinfo{year}{1989}\natexlab{}.
\newblock \emph{\bibinfo{title}{Learning from delayed rewards}}.
\newblock \bibinfo{thesistype}{Ph.D. Dissertation}. \bibinfo{school}{King's
  College}, \bibinfo{address}{Cambridge, United Kingdom}.
\newblock


\end{thebibliography}

\end{document}